\documentclass{article}

\usepackage{times,amsmath,amssymb}
\usepackage{subfigure} 
\usepackage{graphicx} 
\usepackage{url,color} 
\usepackage{booktabs}
\usepackage{natbib}
\usepackage{makecell}
\usepackage{hyperref}

\newtheorem{theorem}{Definition}
\usepackage[accepted]{icml2018}

\begin{document} 

\twocolumn[
\icmltitle{How Robust are Deep  Neural Networks?}
           
\begin{icmlauthorlist}
\icmlauthor{Biswa Sengupta}{hua}
\icmlauthor{Karl J. Friston}{ucl}
\end{icmlauthorlist}

\icmlaffiliation{hua}{AI Theory Lab (Noah's Ark, Huawei), London, UK}
\icmlaffiliation{ucl}{University College London, London, UK}

\icmlcorrespondingauthor{Biswa Sengupta}{biswasengupta@yahoo.com}

\icmlkeywords{recurrent neural networks, pseudo-spectrum}

\vskip 0.3in

]

\printAffiliationsAndNotice{} 



\begin{abstract}
Convolutional and Recurrent, deep neural networks have been successful in machine learning systems for computer vision, reinforcement learning, and other allied fields. However, the robustness of such neural networks is seldom apprised, especially after high classification accuracy has been attained. In this paper, we evaluate the robustness of three recurrent neural networks to tiny perturbations, on three widely used datasets, to argue that high accuracy does not always mean a stable and a robust (to bounded perturbations, adversarial attacks, etc.) system. Especially, normalizing the spectrum of the discrete recurrent network to bound the spectrum (using power method, Rayleigh quotient, etc.) on a unit disk produces stable, albeit highly non-robust neural networks. Furthermore, using the $\epsilon$-pseudo-spectrum, we show that training of recurrent networks, say using gradient-based methods, often result in non-normal matrices that may or may not be diagonalizable.    Therefore, the open problem lies in constructing methods that optimize not only for accuracy but also for the stability and the robustness of the underlying neural network, a criterion that is distinct from the other.   
\end{abstract}


\section{Introduction}

A long-standing issue in the study of recurrent neural networks (RNN) is its inability to retain a memory of long sequences \citep{Bengio1994}. This is because of a fundamental problem of vanishing and exploding gradients, when the spectral norm (i.e., principal eigenvalue) of the recurrent matrix differs from unity. If the spectral norm is greater than 1, gradients grow exponentially during back-propagation, which is known as the exploding gradient problem. Similarly, if the spectral norm is less than 1, the gradient vanishes exponentially, quickly causing catastrophic forgetting. Spectral norm is directly related to the stability of the underlying dynamics wherein eigenvalues outside the unit disk reflect instability of a discrete system. Another issue comprising neural networks is the robustness to tiny perturbation of the weight matrix where unintended or at times adversarial perturbations to the network can result in a structurally different dynamics (structurally unstable) in contrast to an unperturbed system.  Therefore, understanding the relationship between the (eigen) spectrum of the recurrent weights, (dynamical) stability and robustness (structural stability) of a neural network is crucial in the design of safety-critical systems, such as self-driving cars, among others. 

In what follows, we will review several ways to confine the spectral radius of neural networks, to alleviate instability as well as to improve the fidelity of a network to remember long sequences.

\subsection*{Related works}
Since learning the recurrent weight matrix (during training) using back-propagation does not always confine the spectral radius to 1, the simplest solution to this problem is attained by echo state machines \citep{Lukosevicius2009}, that eschew can weight-learning. Echo state machines, with hand-crafted weight matrices, have their largest eigenvalue set to be slightly smaller than 1; therefore, long-term dependencies decay exponentially with time. Other heuristic solutions have concentrated on using a scaled identity matrix to initialize the recurrent weight matrix \citep{Le2015}. Often time, regularizing the recurrent weight matrices using an L1/L2 penalty or using heuristics such as clipping of gradients have proven effective for controlling the spectral radius \citep{Pascanu2013a}. Long Short-Term Memory (LSTM) \citep{Hochreiter1997} avoided some of the problems by introducing additional gates to gauge the flux of information in the recurrent units; another proposition has been to use the second-order geometry  \citep{LeCun2012}. Efforts have also been made towards gradient-free evolutionary optimisation algorithms such as CoDeepNEAT for finessing the neural network architecture \citep{Miikkulainen2017}, without much attention to the stability and the robustness of the weight matrices. Similarly, Neural Turing Machines \citep{Graves2014}, that is an augmented RNN has a (differentiable) external memory that can be selectively read or written to, enabling the network to store the latent structure of longer sequences. Attention networks, on the other hand, enable the RNNs to attend to snippets of their inputs \citep{Vinyals2015}. Other solutions to increase the fidelity for longer sequences has resulted in homotopy continuation of the loss function \citep{Bay2017a}, utilising a stack of recurrent networks \citep{Bay2017b} and learning the local information geometry of the recurrent encoding \citep{Bay2017c}. 

A large proportion of work has concentrated upon controlling the gain of the weight matrix $W$. This is done by ensuring that $W$ is close to an orthogonal matrix, by factorizing it according to its singular value decomposition $USV^{T}$ with $U$ and $V$ as the orthogonal bases and $S$ a diagonal matrix containing the singular values. Such a factorization gives us a convenient way to bound the spectral norm (maximum gain) as well as the contractivity (minimum gain) of the matrix $W$. Unitary RNNs \citep{Arjovsky2016} have been proposed to learn a unitary weight matrix with eigenvalues constrained at 1. Later, using Sard's theorem, it was shown that if unitary matrices are parameterised by $p$ real-valued parameters, they cannot represent the entire set of unitary matrices \citep{Wisdom2016}. The underlying Stiefel manifold, on the other hand, enables a geodesic descent (via Cayley transformation) and imposes $U^{T}U=I$ and ${V^{T}V=I}$, during the descent. This lifts the capacity constraints displayed by the original unitary RNNs. Related work also uses the Cayley transformation \citep{Helfrich2017} to ensure orthogonality. Other attempts to enforce orthogonality constraints have resulted in a soft constraint on the diagonal matrix $S$  directly \citep{Vorontsov2017}. \citet{Mhammedi2016}, on the other hand, have used a complex, unitary transition matrix to pose a new RNN. Specifically, they map the weight matrix using Householder reflections to guarantee orthogonality. 


In summary, these methods approach the problem of avoiding catastrophic forgetting of long sequences quite differently. They include direct solutions such as architectural modifications, gradient-free/homotopy methodologies and imposing structure on recurrent connections (orthogonality), as well as indirect solutions such as imposing structure on the latent space, initialisation strategies, using the 2nd order geometry, etc. However, all of these studies concentrate on the remembering/forgetting of longer sequences, addressing the issue of stability but ignoring robustness criteria that are critical in the design of safe-AI solutions. Moreover, it is always assumed that training a recurrent neural network always results in a stable system -- the underlying hypothesis, therefore, tells that a highly accurate neural network is always stable. Is it really the case?

\subsection*{Our contribution}
The contribution of this work includes:

\begin{enumerate}
\item Illustrate on three benchmark tasks, the instability of learned recurrent matrices of an RNN, an LSTM and a GRU based recurrent network.
\item Illustrate that the learned weight matrices of RNN, LSTM and GRU can often be non-normal.
\item Illustrate that spectral normalization, to constrain eigenvalues inside the unit disk, is a first step towards stabilizing the discrete dynamics, however, it does not address the issue of robustness.
\end{enumerate}




\section{Methods}
\label{sec:methods}

In this section, we will describe the terminology that we use in our results section. Specifically, we will go through the following steps to illustrate non-normality of an RNN: first, for 3 different benchmark problems, we will train an RNN, an LSTM and a GRU. We will then use the learned weight matrices to compute Henrici's number, a quantitative measure of non-normality and pseudo-spectrum, a qualitative illustration of the robustness of the discrete dynamical system (here, a neural network). After normalizing the spectrum using the Power method \citep{Golub1996}, we will again evaluate how stability and robustness are affected. 








\subsection*{Mathematical Statement}

Three types of recurrent neural networks -- RNN, LSTM and GRU, modelled as per \cite{Goodfellow2016} are used for our analysis.

Assume an RNN, with input $u$ and state $x$, where a loss function $\mathcal{L}$ is minimized iteratively as

\begin{eqnarray}
  {x_t} & = & W\sigma \left( {{x_{t - 1}}} \right) + {W_{in}}{u_t} + b \hfill \nonumber \\
  \mathcal{L} & = & \sum\limits_{t = 1}^T {{\mathcal{L}_t}\left( {{x_t}} \right)}  \hfill \nonumber \\ 
\end{eqnarray}

$W$ is the recurrent weight matrix, ${W_{in}}$ is the input weight matrix and $b$ is the bias term. The gradient descent to learn parameters $\left\{ {W,{W_{in}},b} \right\} \in \theta$ can be instantiated using the chain rule as,

\begin{eqnarray}
  \frac{{\partial L}}{{\partial \theta }} & = & \sum\limits_{t = 1}^T {\frac{{\partial {L_t}\left( {{x_t}} \right)}}{{\partial \theta }}}  \hfill \nonumber \\
  \frac{{\partial {L_t}\left( {{x_t}} \right)}}{{\partial \theta }} & = & \sum\limits_{1 \leqslant k \leqslant t} {\left( {\frac{{\partial {L_t}}}{{\partial {x_t}}}\frac{{\partial {x_t}}}{{\partial {x_k}}}\frac{{\partial {x_k}}}{{\partial \theta }}} \right)}  \hfill \nonumber \\
  \frac{{\partial {x_t}}}{{\partial {x_k}}} & = & \prod\limits_{t \geqslant i > k} {\frac{{\partial {x_i}}}{{\partial {x_{i - 1}}}}}  = \prod\limits_{t \geqslant i > k} {{W^T}} \left( {\sigma '\left( {{x_{i - 1}}} \right)\mathbf{I}} \right) \hfill  \nonumber \\
\end{eqnarray}

It is then convenient to see that the 2-norm of the Jacobian (${\frac{{\partial {x_t}}}{{\partial {x_k}}}}$) is bounded by the product of norms of the recurrent weight matrix $W$ and that of the non-linearity (${\sigma '\left( {{x_{i - 1}}} \right)\mathbf{I}}$). For a linear neural network, therefore, the convergence of the gradient descent is directly related to the spectral norm of the recurrent weights. Specifically, forcing the spectral norm of $W$ to 1 alleviates vanishing or exploding gradients.

In the work surveyed above, it is assumed that the recurrent matrix is normal. A matrix $W$ is normal if it commutes with its adjoints i.e., ${W^ * }W = W{W^ * }$. Alternatively, ${W}$ is normal if and only if it is unitarily diagonalizable i.e., ${W=USV^{*}}$ s.t. ${U{V^ * } = I}$ and ${S}$ is a diagonal matrix, else it is known as a defective matrix. The extent of non-normality can be computed via Henrici's number i.e., $\nu=\frac{\left \| WW^{*} -W^{*}W \right \|}{\left \| W^{2}  \right \|}$. A ``large'' value indicates spectral instability while 0 indicates normality. We will use Henrici's number to evaluate the non-normality of the different classes of learned weight matrices.

In order to qualitatively evaluate robustness, we make use of the $\epsilon$-pseudospectrum. Put simply, it tells us how the spectral portrait of an operator changes as we perturb it. Formally, $\epsilon$-pseudospectrum can be defined as,

\begin{theorem}
Let $W$ be a square n-by-n matrix of complex numbers. For $\epsilon>0$, the $\epsilon$-pseudospectrum of $W$ is defined as,

\[{\sigma _\varepsilon }\left( W \right) = \left\{ {\lambda  \in \mathbb{C}:\left\| {{{\left( {W - \lambda I} \right)}^{ - 1}}} \right\| \geqslant {\varepsilon ^{ - 1}}} \right\}\]
\end{theorem}

Using Kreiss' theorem \citep{Golub1996}, we can now also bound the pseudospectrum as a function of exponent of the matrix i.e., $||W^{l}||$ can be bounded as,

\begin{eqnarray}
\mathop {\sup }\limits_{\varepsilon  > 0} \frac{{{\rho _\varepsilon }(W) - 1}}{\varepsilon } \le \mathop {\sup }\limits_{l \ge 0} \left\| {{W^l}} \right\| \le {\rm{e \ n }}\mathop {\sup }\limits_{\varepsilon  > 0} \frac{{{\rho _\varepsilon }(W) - 1}}{\varepsilon }
\end{eqnarray}

where $l$ is the number of layers (coefficient of matrix powers) and $n$ is the dimension of the matrix. We will numerically calculate the pseudospectrum to evaluate the robustness of the weight matrix in the later sections.

\subsubsection*{Block-diagonals of Schur Decomposition}

Theoretical work suggests that ‘weight’ matrices in neuronal (i.e., biological) networks can be non-normal \citep{Ganguli2008,Murphy2009,Goldman2009}. Often times, for a non-normal matrix (identified using the pseudo spectrum or Henrici's number) a set of linearly independent eigenvectors may not exist, or they may well be far from being orthogonal.  


In order to factor such non-normal matrices we make use of the Schur decomposition (Schur's Unitary Triangularization Theorem) of the matrix such that $W = Q\left( {\lambda \mathbb{I} + N} \right){Q^ * }$. Here, $\mathbb{I}$ is the identity matrix, $\lambda$ are the eigenvalues, $Q$ is a unitary matrix and $N$ is a strictly upper triangular matrix. Notice that the non-normal part $N$ is not unique, although ${{{\left\| N \right\|}_F}}$ is. The volume of $N$ is yet another indicator for non-normality.


\subsubsection*{Spectral stabilization}
A discrete dynamical system is unstable if the spectral radius is outside the unit disk. Often times, the weight matrices that are learnt by a back-propagation algorithm can become unstable. Spectral instability implies non-normality, however non-normality does not imply instability. A straight-forward method to constrain the spectrum (without assuming non-normality)  inside the unit disk can be formulated using the power method \citep{Golub1996} i.e., by finding the dominant eigenvalue and normalizing the other eigenvalues as,


\begin{algorithm}[tb]
   \caption{Spectrum Stabilization}
   \label{algo:ss}
\begin{algorithmic}
   \STATE {\bfseries Input:} raw weight matrix $W$, no. of power iterations $m$
    \STATE {\bfseries Output:} stabilized weight matrix $W_{s}$
   \STATE $u_0 \sim \mathcal{N}\left( {0,\sigma } \right)$
   \FOR{$i=1$ {\bfseries to} $m$}
   \STATE $v_{i} = \frac{{{W^T}u_{i-1}}}{{{{\left\| {{W^T}u_{i-1}} \right\|}_2}}}$
   \STATE $u_{i} = \frac{{{W}v_{i}}}{{{{\left\| {{W}v_{i}} \right\|}_2}}}$
   \ENDFOR
   \STATE $W_{s} = \frac{W}{{u_{{{\rm{i}}_{{\rm{max}}}}}^TWv_{{{\rm{i}}_{{\rm{max}}}}}^{}}}$
\end{algorithmic}
\end{algorithm}

For our experiments $i_{max}$ is fixed at 1.
\subsection*{Datasets}

In this section, we describe the three benchmark examples that we have used to demonstrate the non-normality of the weight matrices obtained post-training:

\begin{itemize}
\item Task 1 -- Adding two numbers \citep{Hochreiter1997}: This task is the simplest of the three tasks that we have studied in this paper. It comprises the learning of long-term dependencies required to add two numbers – after the recurrent networks have been trained on a randomly generated tuple of two numbers and their sum (training size = 45,000; test size = 5,000). Parameters for simulation involving RNN, LSTMS and GRU include -- batch size: 128 and hidden units: 128. The weights were randomly initialized. 

\item Task 2 -- Sentiment analysis using the IMDB dataset \citep{Maas2011}: Here, reviews about individual movies from the IMDB data-set are represented as a variable sequence input of words and the sentiment of the review forms an output. The dataset contains 25,000 movie reviews for training and the same number are available for testing. Prior to training a recurrent network, each movie review is embedded in a latent space where the similarity between words in terms of meaning is translated to closeness in a vector space. The number of hidden layers in the RNN and LSTM/GRU was set to 16 and 32, respectively.

\item Task 3 -- Sequential MNIST task \citep{Le2015}: This dataset entails classification of MNIST images to 10 classes, where each image has a dimension of $28 \times 28$ pixels. Using the RNNs, we consider each MNIST image as a sequence of length 28 (number of image rows) with a 28-dimensional input vector i.e., each image row has 28 columns associated with a single time step. Parameters for simulation involving RNN, LSTMS and GRU include -- batch size: 32, epochs: 200, learning rate: 1e-6 and hidden units: 100. The weights were randomly initialized. 
\end{itemize}

We have avoided using any weight initialization, such as setting them to an identity matrix (instead of a random matrix), which has proven to be beneficial in some instance \citep{Le2015}. This was done to retain inference difficulty associated with training a general class of recurrent networks. In the next section, we will examine non-normality of three neural networks -- RNN, LSTM and GRU -- for three different tasks i.e., the addition of two numbers, pixel-wise MNIST, and the IMDB sentiment analysis dataset.

\section{Results}

\begin{figure}[!h]
\centering     
\subfigure[2 layer RNN]{\label{fig:addition_a}\includegraphics[width=40mm]{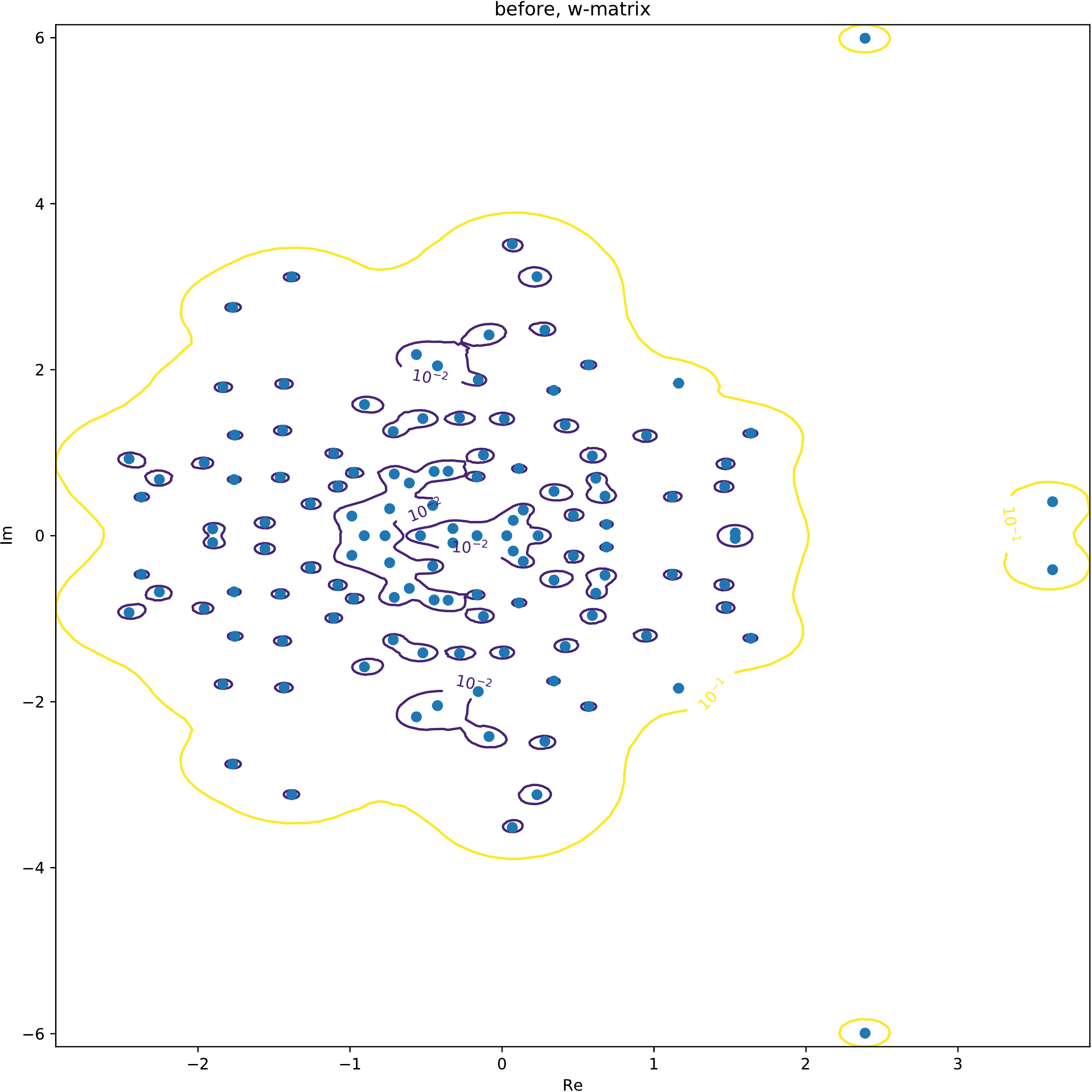}}
\subfigure[2 layer LSTM]{\label{fig:addition_b}\includegraphics[width=60mm]{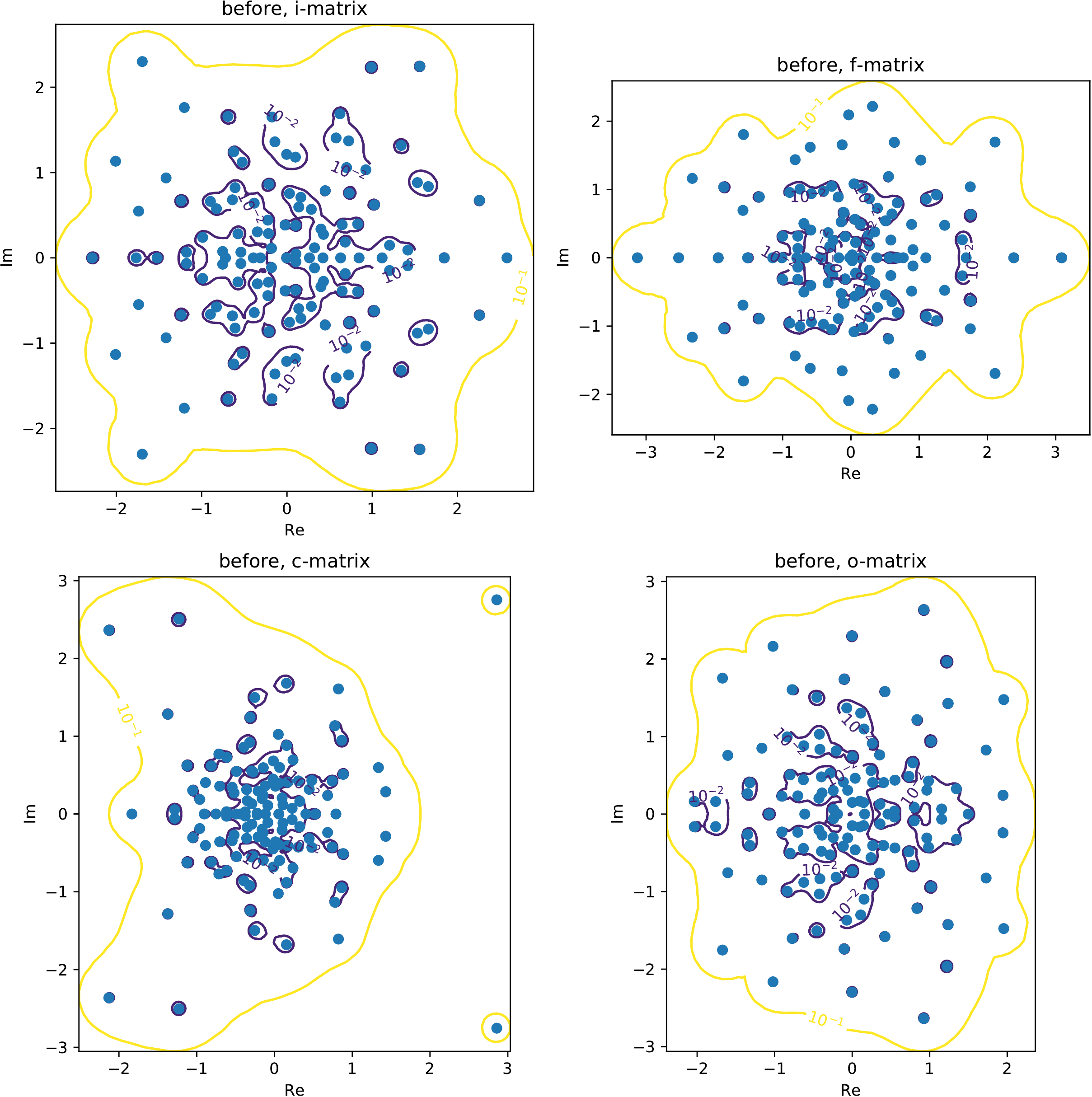}}
\subfigure[2 layer GRU]{\label{fig:addition_c}\includegraphics[width=60mm]{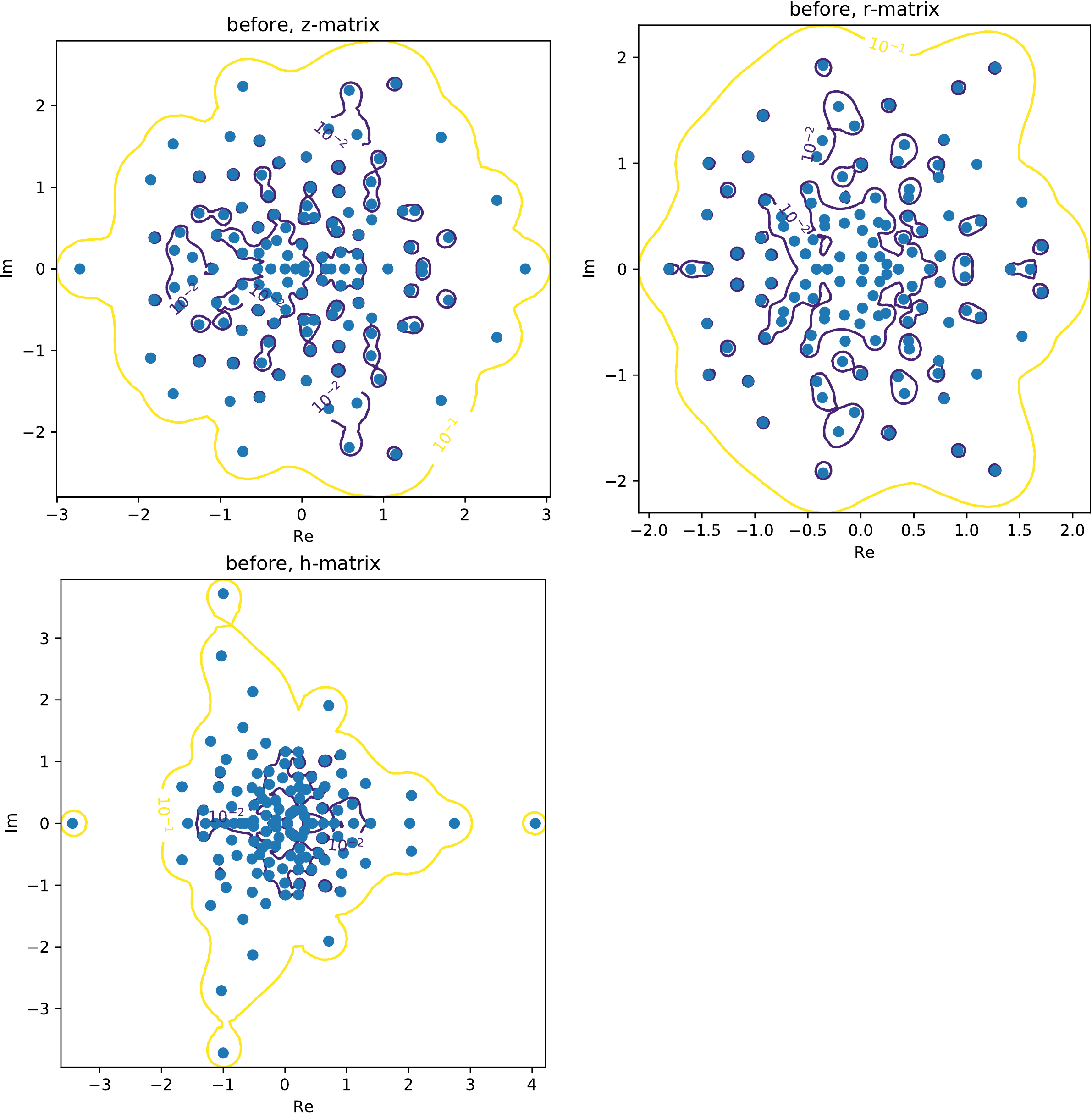}}
\caption{Pseudo-spectrum of an RNN, a LSTM and a GRU for the addition task. The architecture consisted of 2 layers of recurrent networks (the spectrum of only one weight matrix is shown).}
\label{fig:addition}
\end{figure}

\begin{figure}[!h]
\centering     
\subfigure[1 layer RNN]{\label{fig:sentiment_a}\includegraphics[width=40mm]{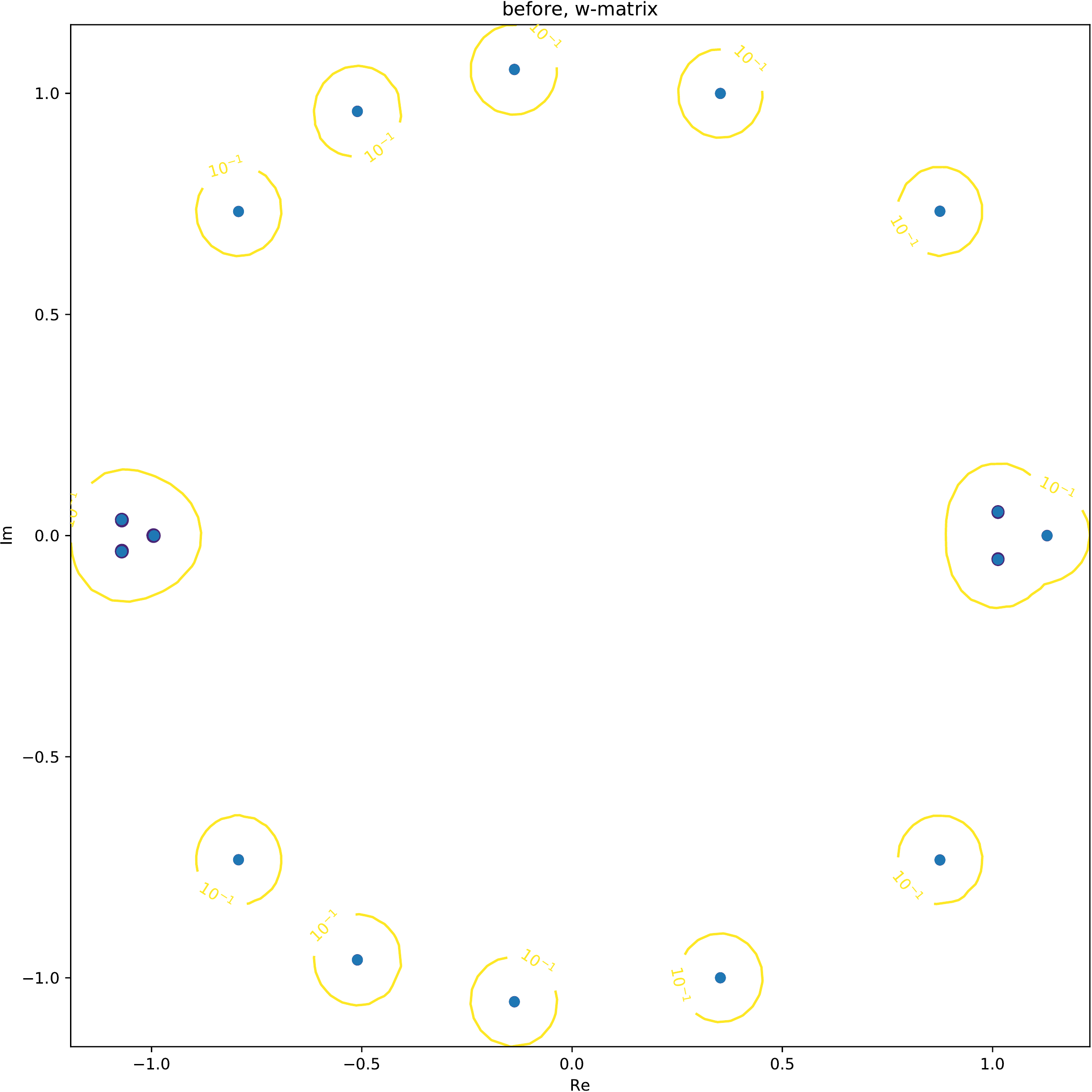}}
\subfigure[1 layer LSTM]{\label{fig:sentiment_b}\includegraphics[width=80mm]{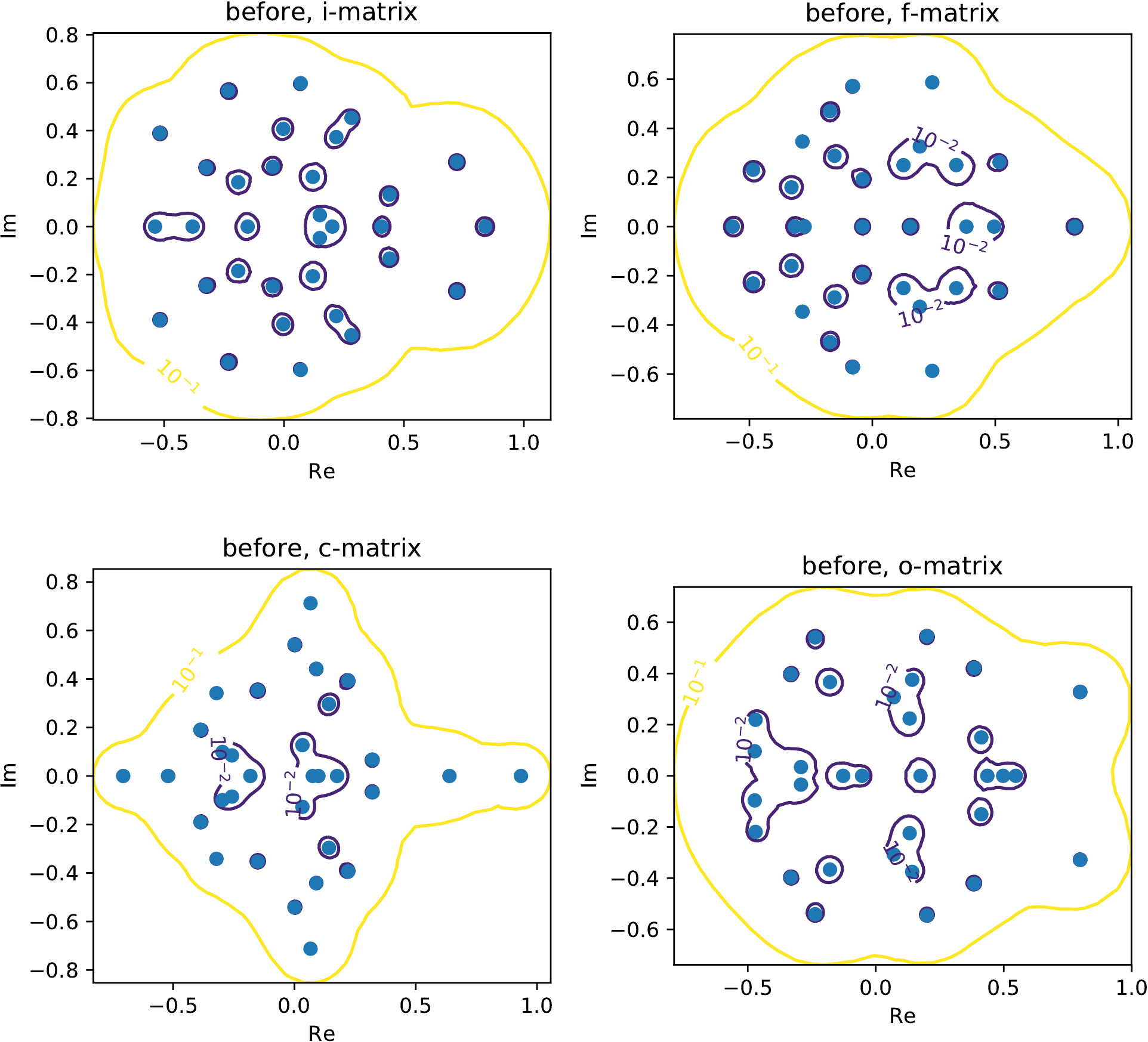}}
\subfigure[1 layer GRU]{\label{fig:sentiment_c}\includegraphics[width=80mm]{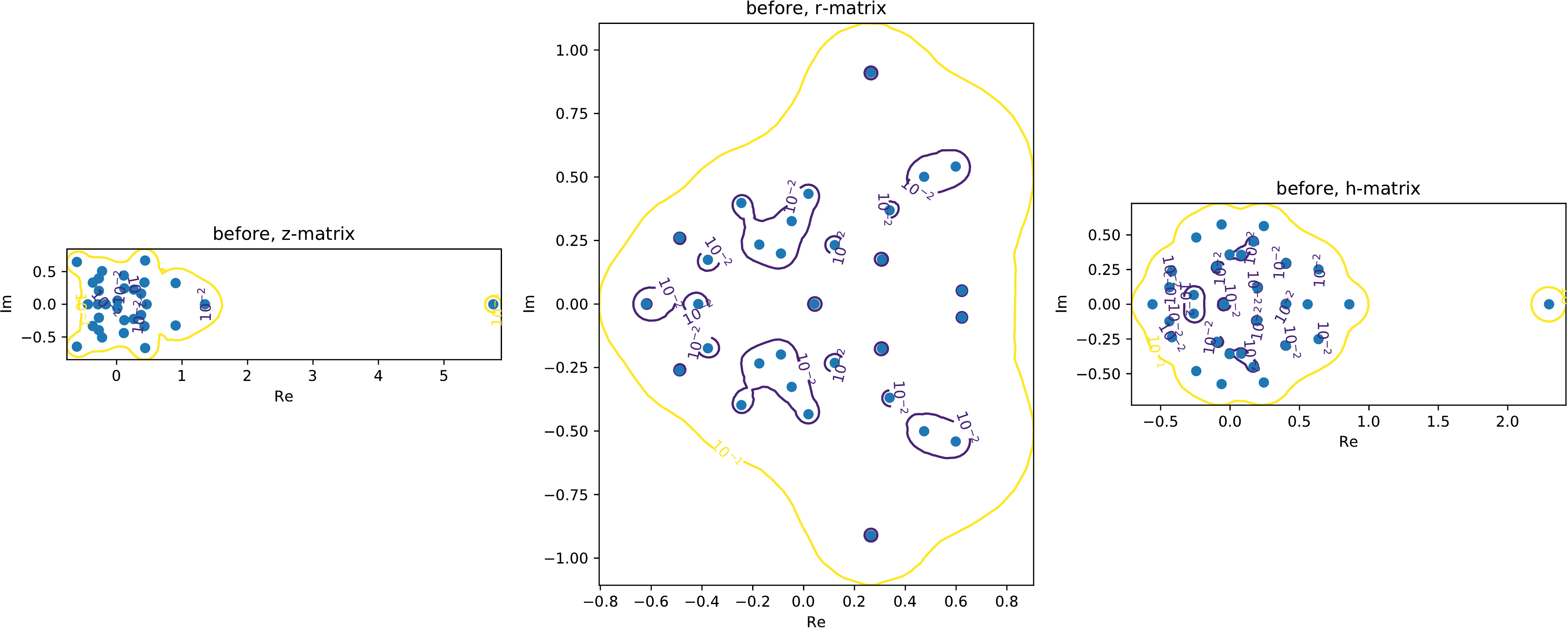}}
\caption{Pseudo-spectrum of an RNN, a LSTM and a GRU for the IMDB sentiment-analysis task.} 
\label{fig:sentiment}
\end{figure}

\begin{figure}[!h]
\centering     
\subfigure[1 layer RNN]{\label{fig:mnist_a}\includegraphics[width=40mm]{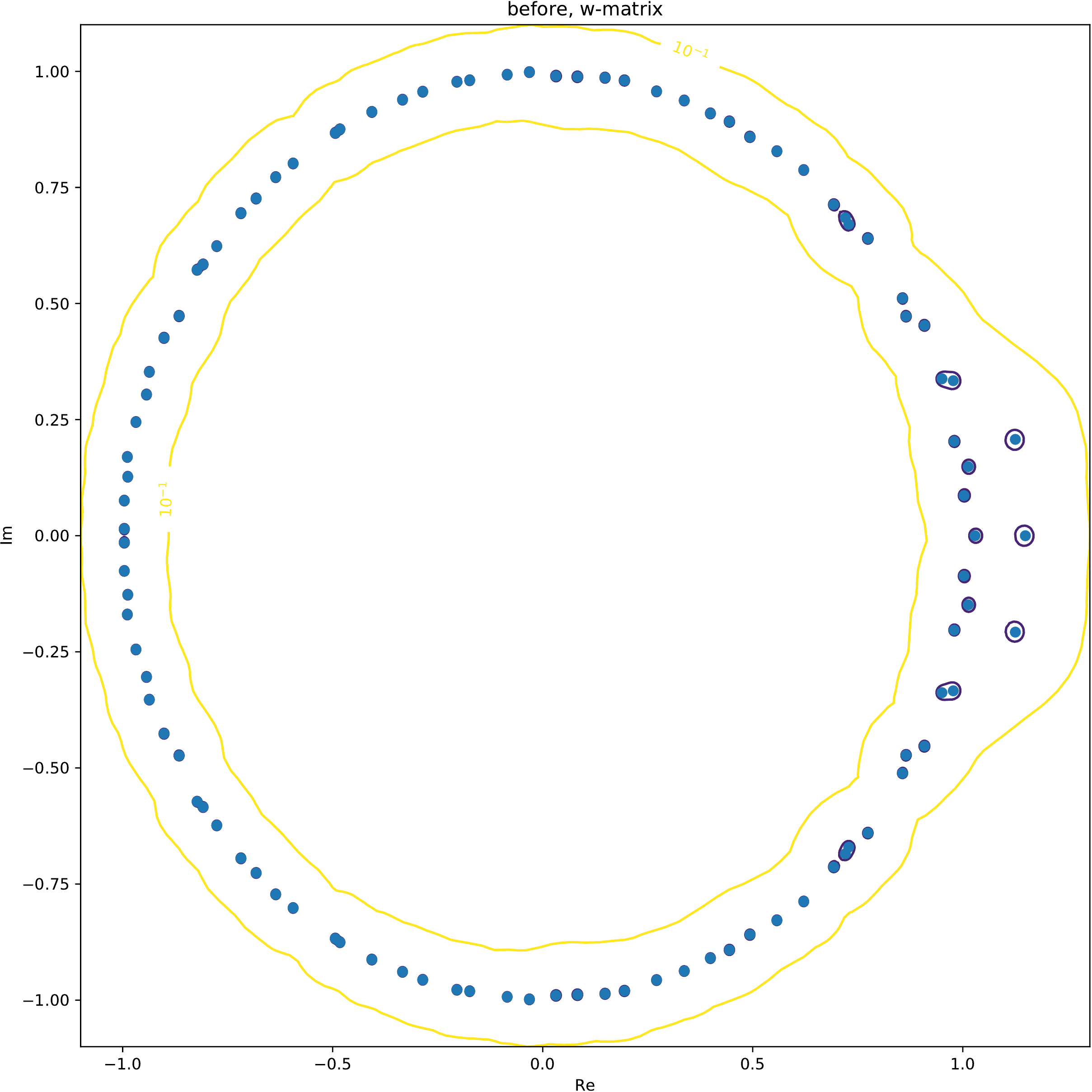}}
\subfigure[1 layer GRU]{\label{fig:mnist_b}\includegraphics[width=60mm]{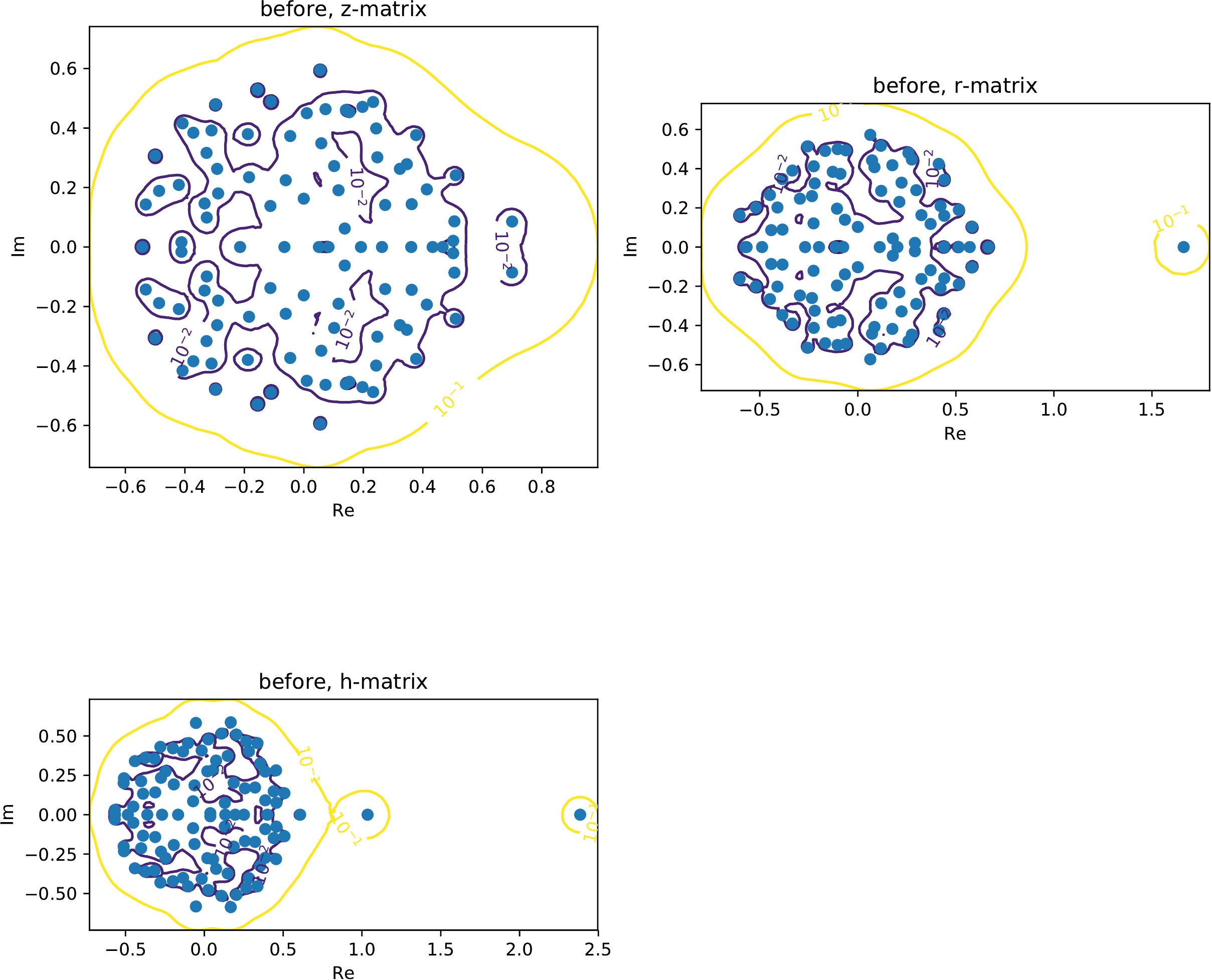}}
\caption{Pseudo-spectrum of an RNN and a GRU for the pixel based MNIST classification task. The LSTM used for the simulation did not converge.} 
\label{fig:mnist}
\end{figure}

Three benchmark problems -- addition, sentiment analysis and pixel-based MNIST were evaluated as described in Section \ref{sec:methods}. Table \ref{table:accuracy_rnn} displays the test accuracies for the different datasets using RNN, LSTM and GRU based neural networks. The three networks applied to adding two numbers did not show statistically significant differences in accuracies; the accuracy of sentiment analysis task was also fairly similar for the three networks. For the pixel-based classification the LSTM model does not converge for three sets of random initialization, therefore this set was not analysed further. Interestingly, initializing weight matrices using an identity matrix \citep{Le2015} causes the accuracy to rise beyond 92\%.

In Table \ref{table:nonnormal_rnn}, we calculate the Henrici number (see Methods) to gauge the extent of non-normality that different learned weight matrices inherit. Of concern, is the weight matrix between the hidden recurrent units that, after training, appear non-normal. This means a tiny perturbation of the weight matrix can completely alter the spectral signature of the underlying recurrent network, making these networks non-robust.

In Figures \ref{fig:addition}, \ref{fig:sentiment} and \ref{fig:mnist} we plot the pseudo-spectrum of the weight matrices. There are two salient issues -- first, notice that although the accuracies for Task 1 and 2 are reasonable, the learned weight matrices have few eigenmodes that are unstable; second, notice the (yellow) contours for task 2 -- they indicate that a tiny perturbation on the weight matrix can cause them to become unstable, pointing towards transient growth in few layers. What this tells is that although the eigen-analysis may suggest stability, there may be few layers where the transient growth can cause instability.

\begin{figure}[!h]
\centering     
\subfigure[2 layer RNN]{\label{fig:addition_a}\includegraphics[width=40mm]{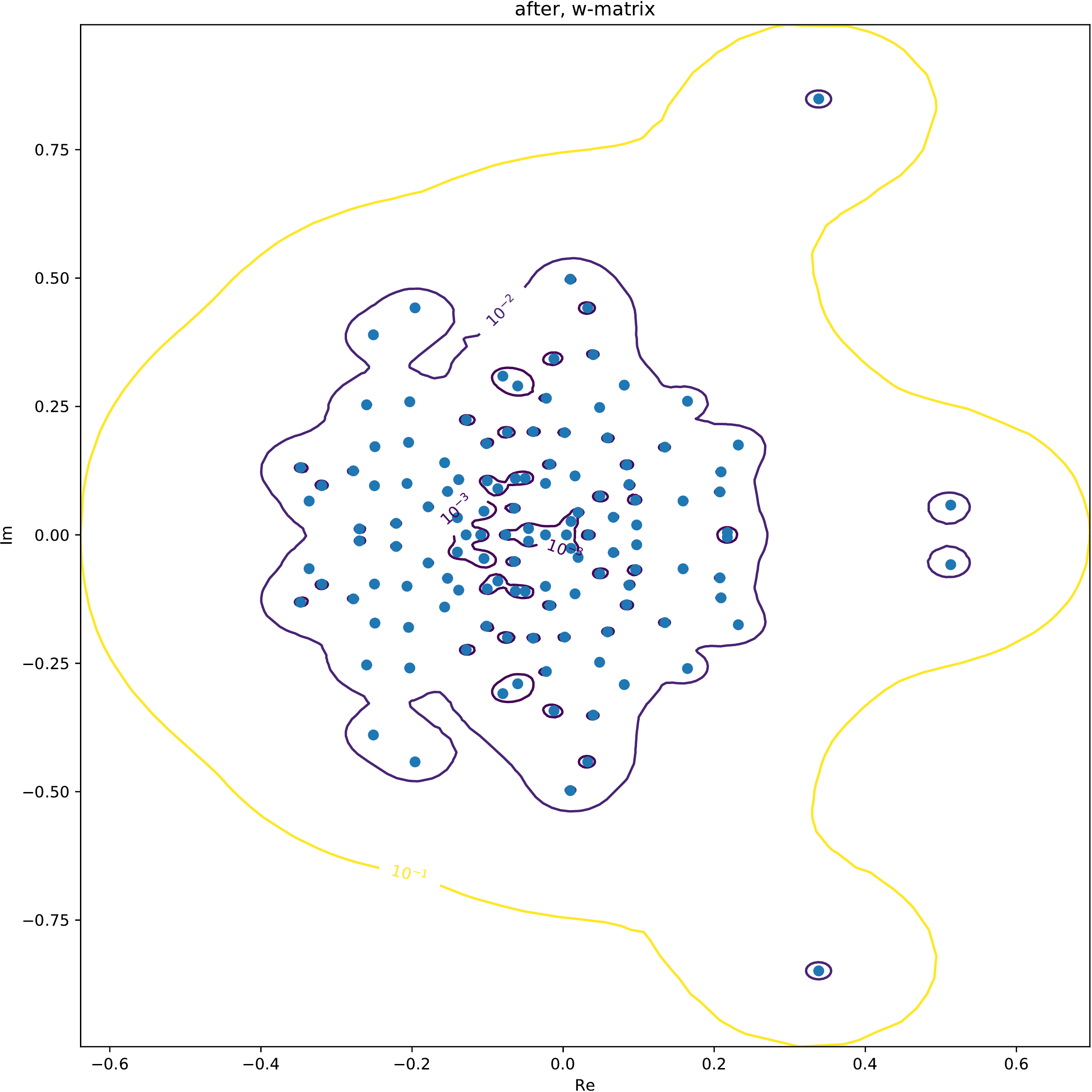}}
\subfigure[2 layer LSTM]{\label{fig:addition_b}\includegraphics[width=40mm]{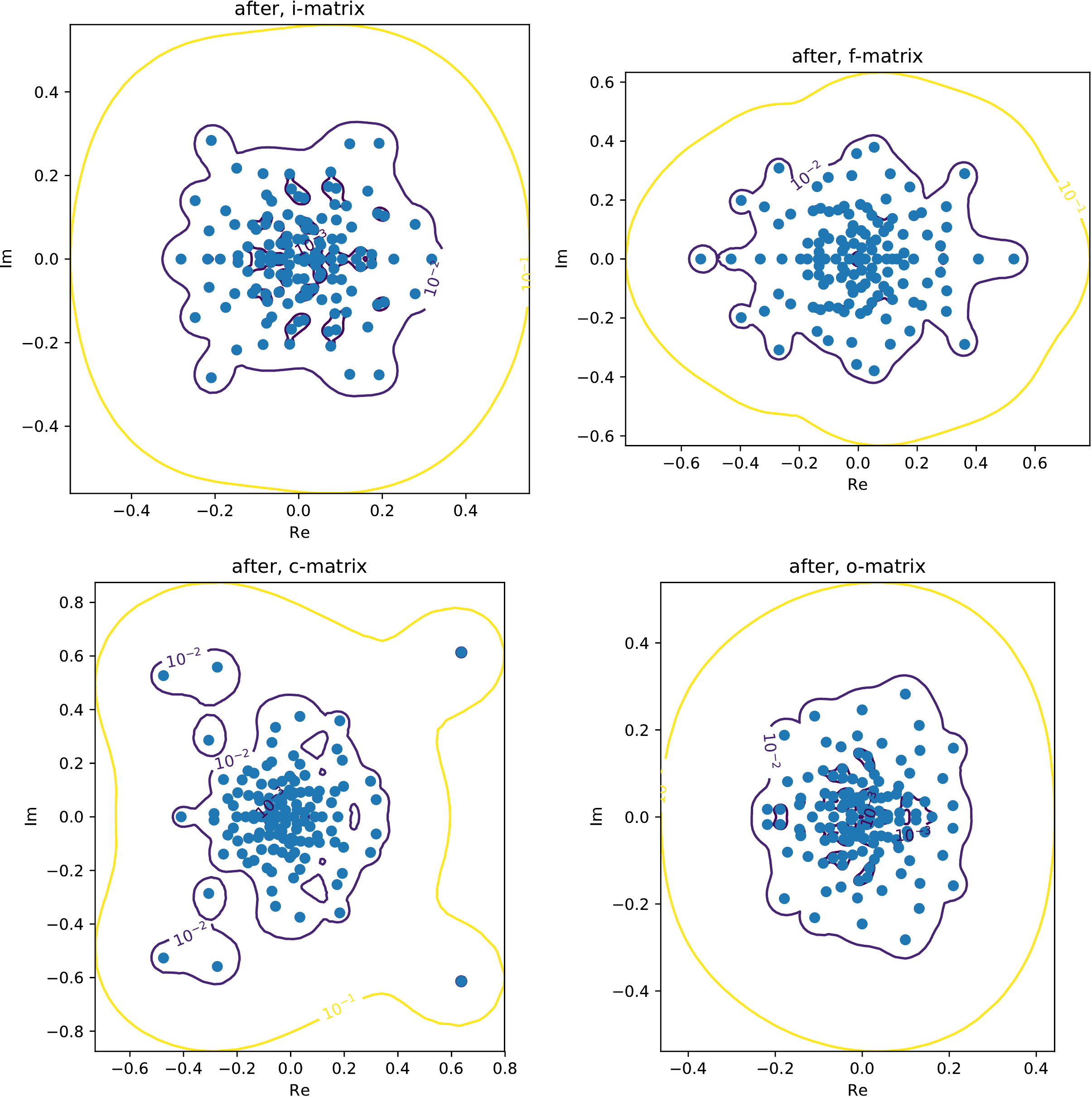}}
\subfigure[2 layer GRU]{\label{fig:addition_c}\includegraphics[width=40mm]{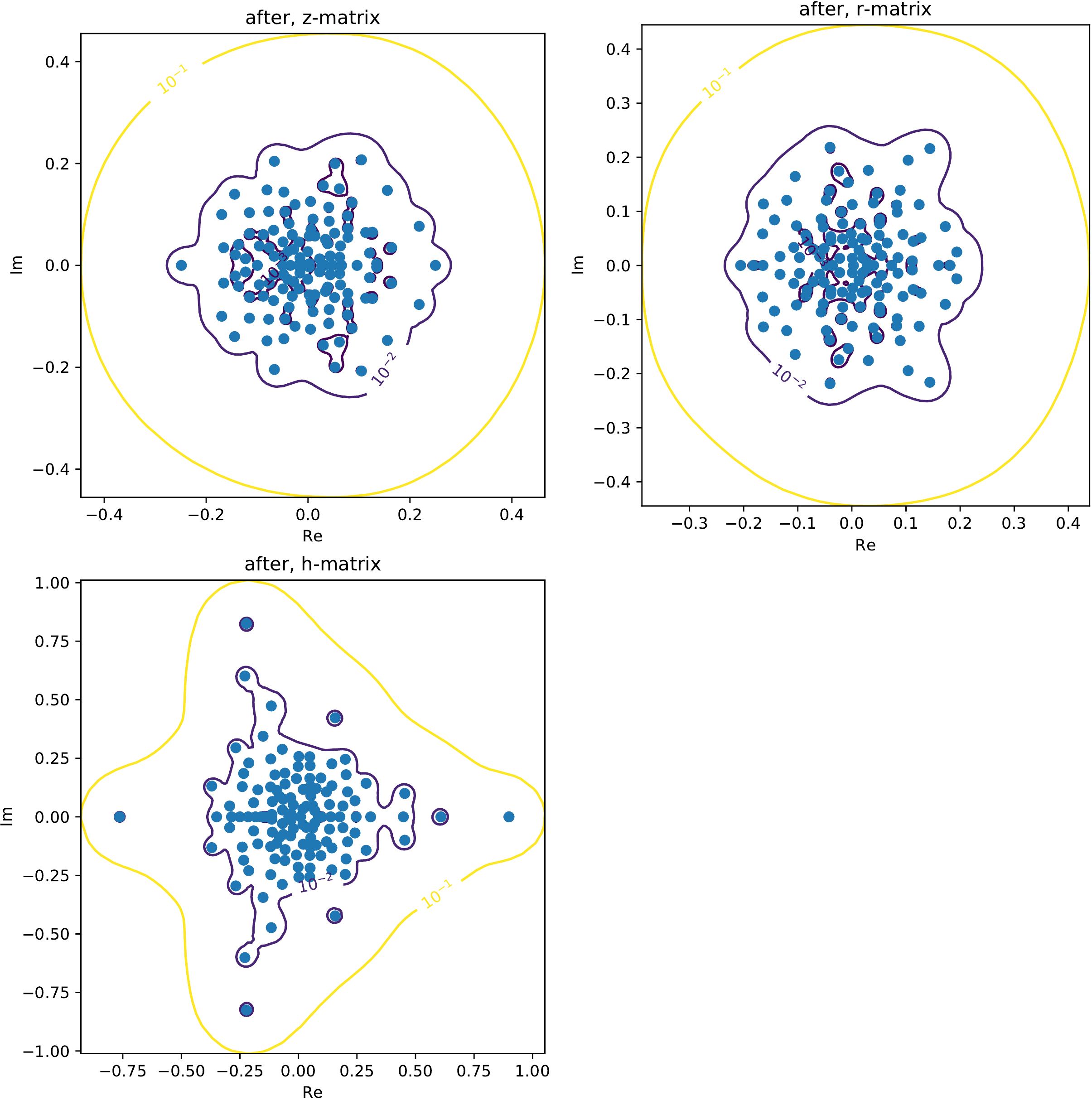}}
\caption{Pseudo-spectrum of an RNN, a LSTM and a GRU for the addition task, after normalizing the spectrum. Stability is regained however robustness is lost.}
\label{fig:addition_after}
\end{figure}

\begin{figure}[!h]
\centering     
\subfigure[2 layer RNN]{\label{fig:sentiment_after_a}\includegraphics[width=40mm]{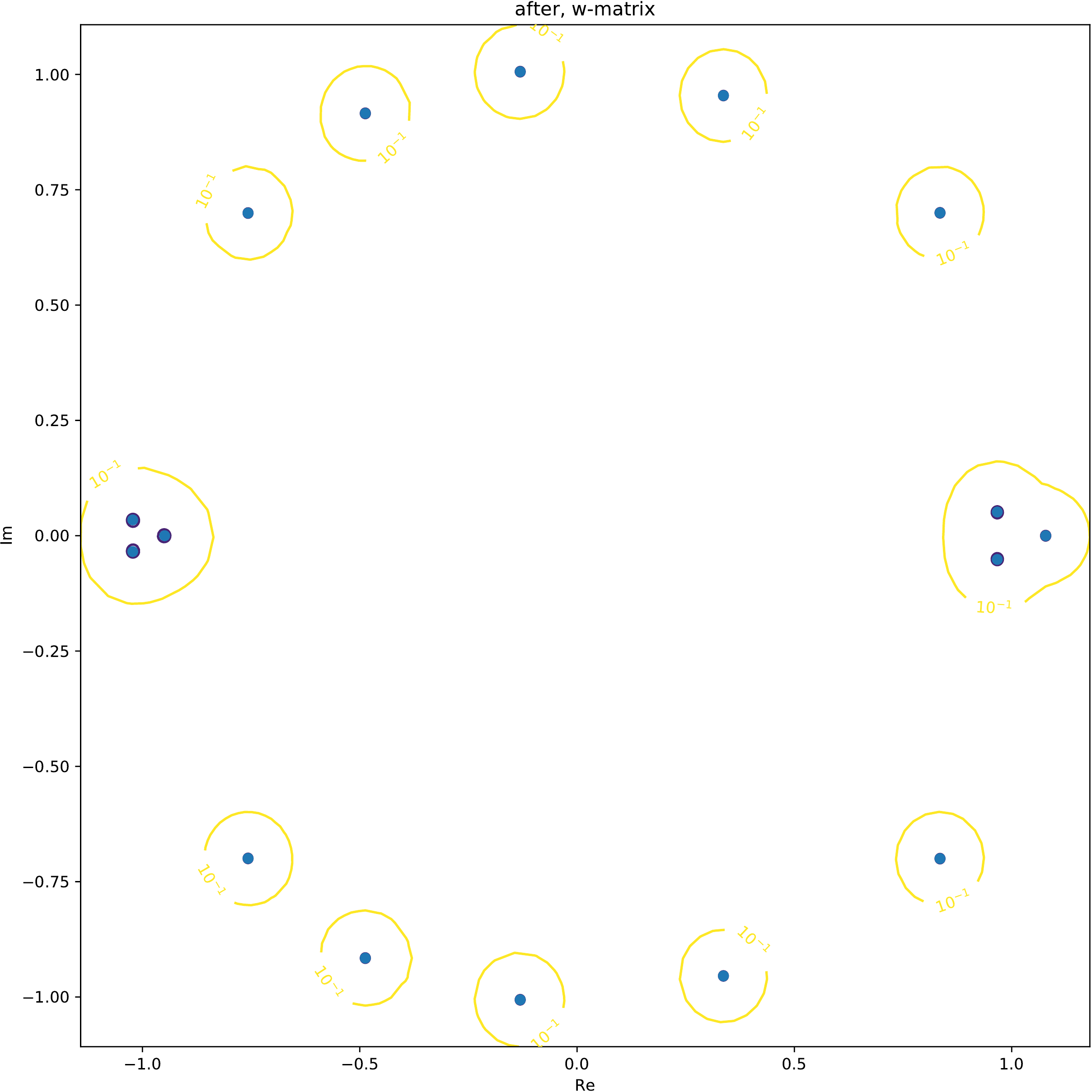}}
\subfigure[2 layer LSTM]{\label{fig:sentiment_after_b}\includegraphics[width=60mm]{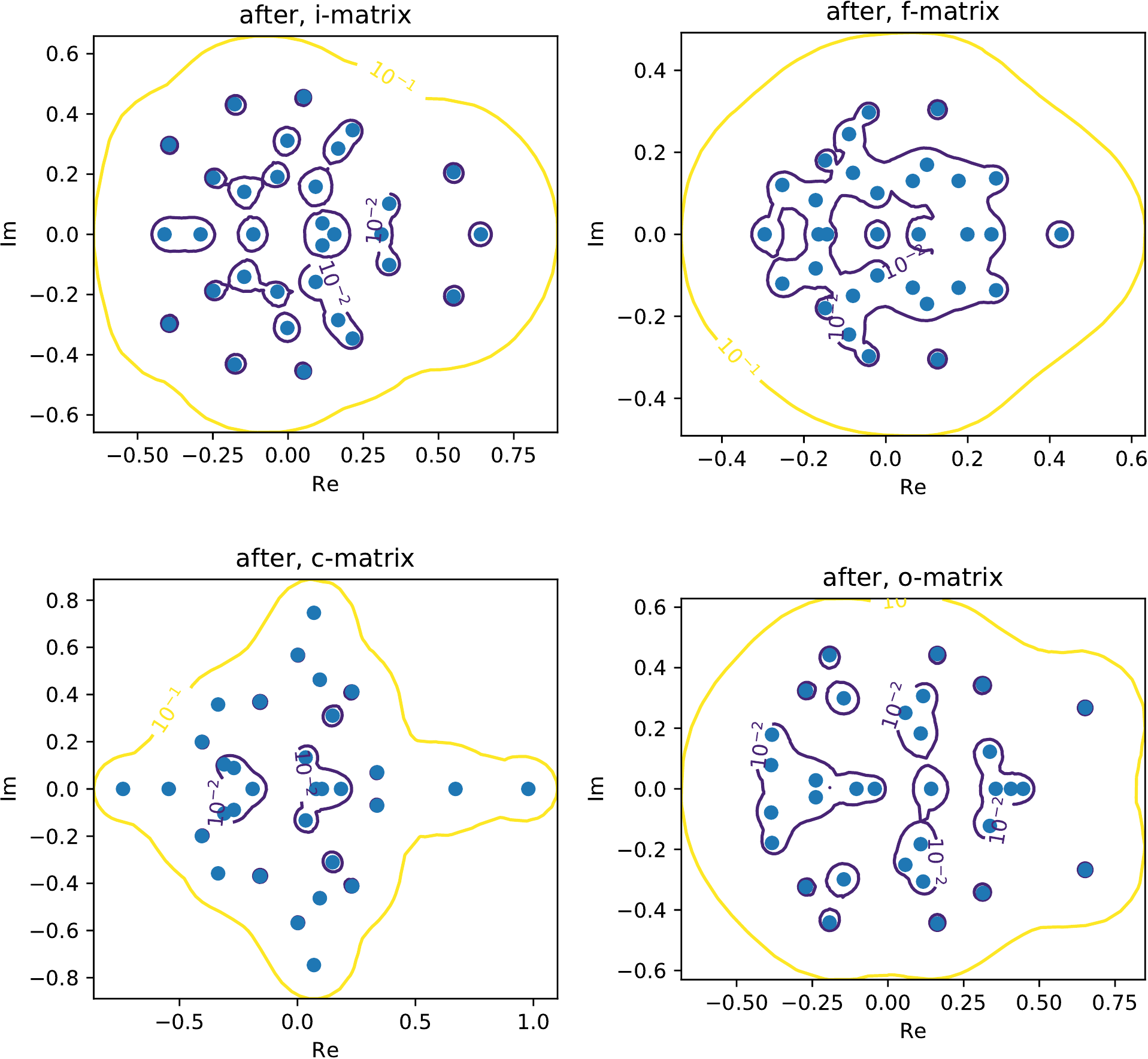}}
\subfigure[2 layer GRU]{\label{fig:sentiment_after_c}\includegraphics[width=80mm]{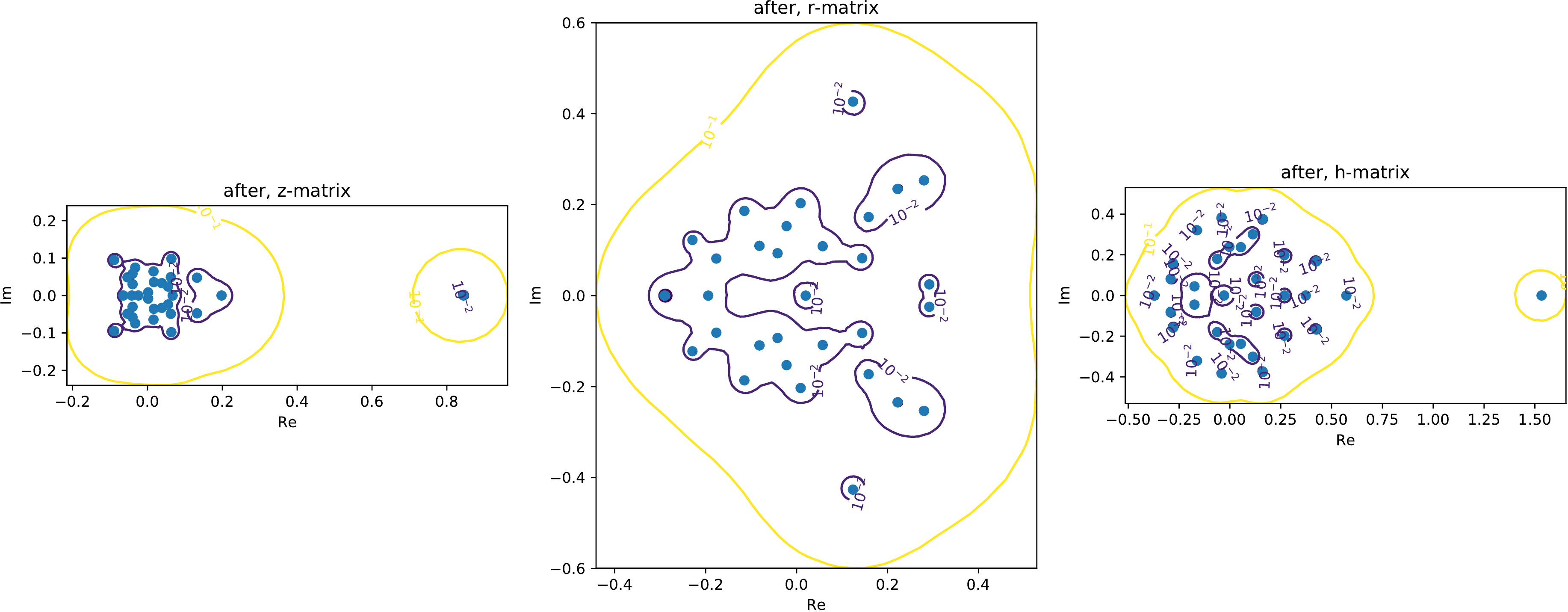}}
\caption{Pseudo-spectrum of an RNN, a LSTM and a GRU for the sentiment-analysis task, after normalizing the spectrum. }
\label{fig:sentiment_after}
\end{figure}

\begin{figure}[!h]
\centering     
\subfigure[1 layer RNN]{\label{fig:mnist_after_a}\includegraphics[width=40mm]{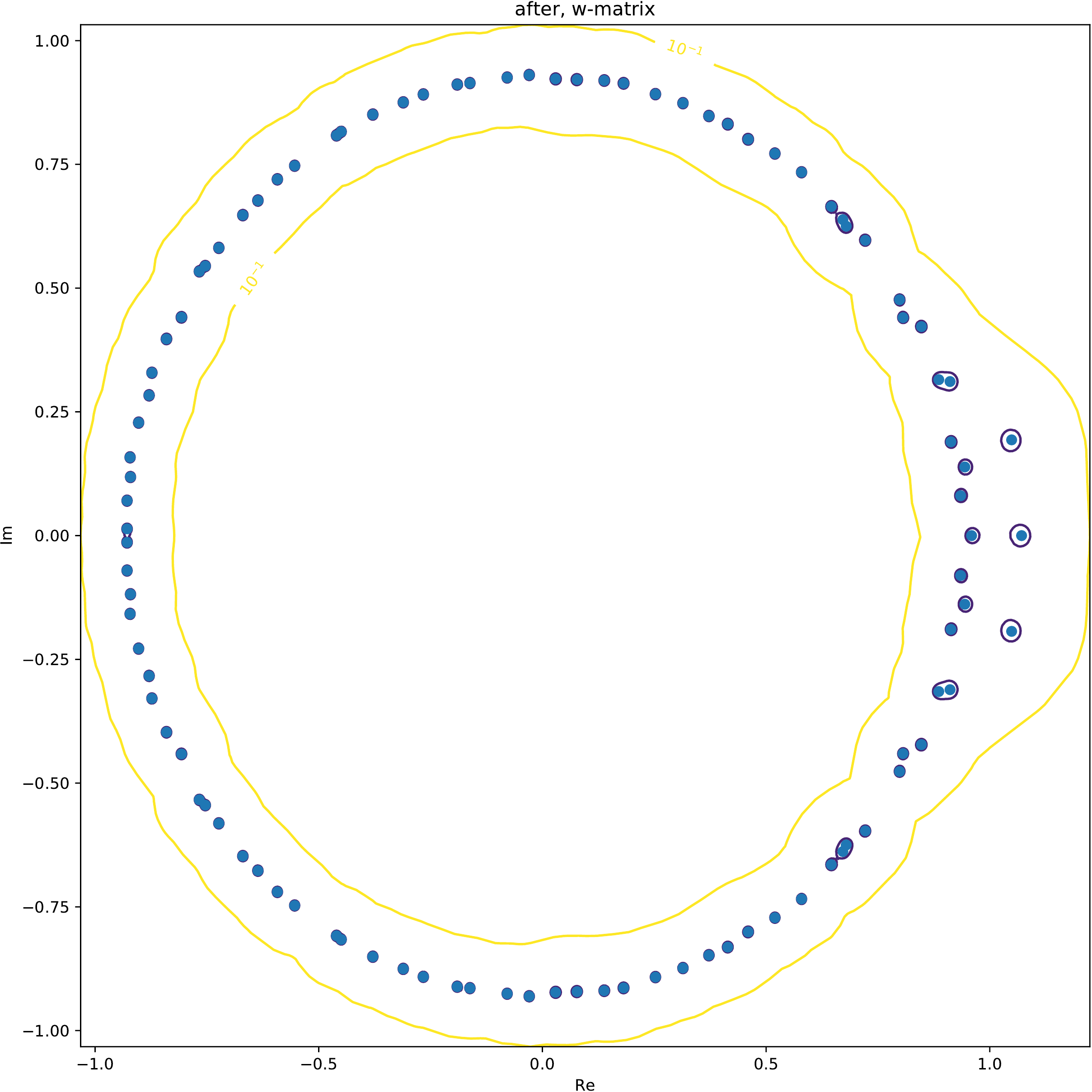}}
\subfigure[1 layer GRU]{\label{fig:mnist_after_b}\includegraphics[width=60mm]{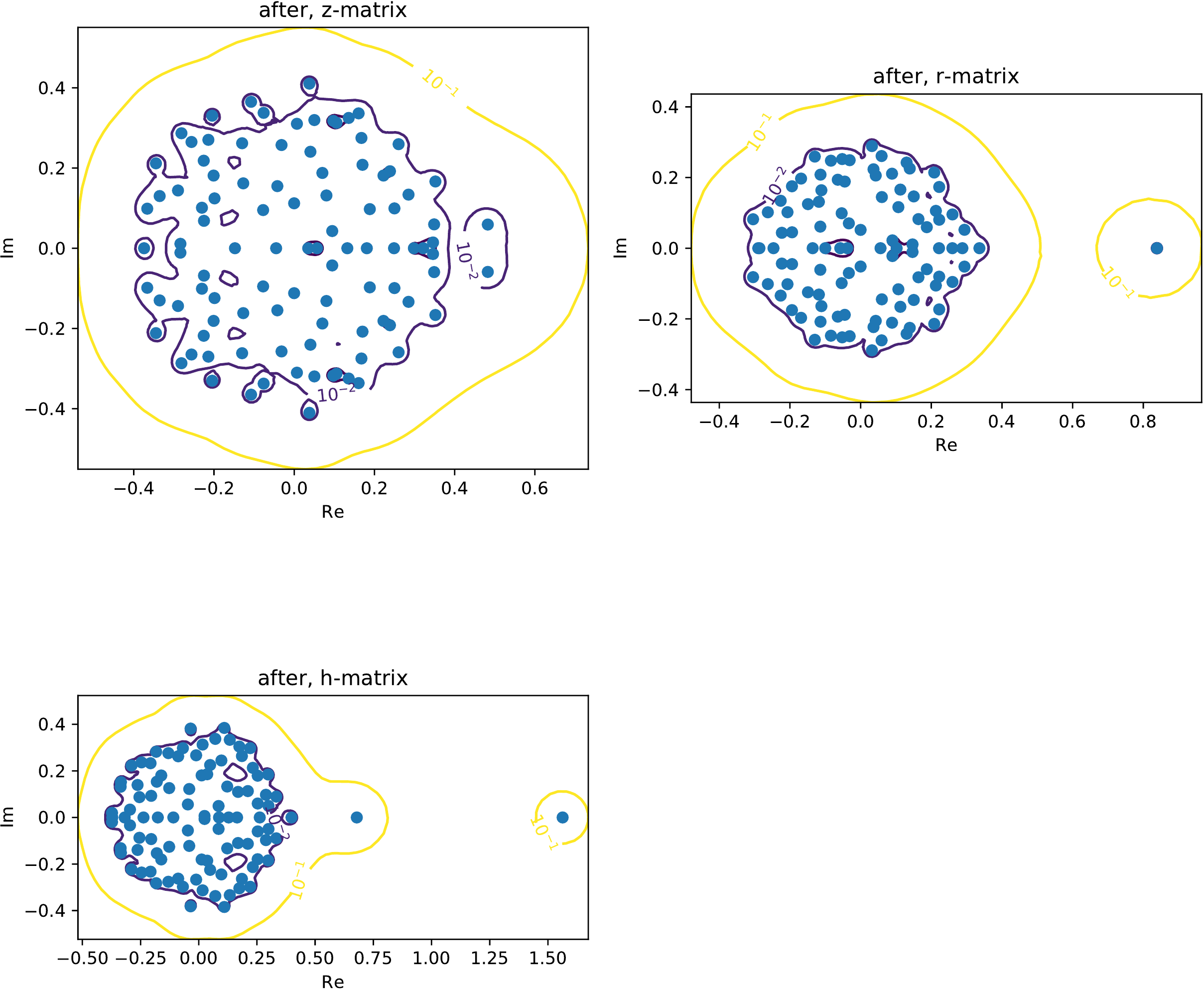}}
\caption{Pseudo-spectrum of an RNN and a GRU for the pixel based MNIST classification task, after normalizing the spectrum.} 
\label{fig:mnist_after}
\end{figure}

One can use spectral stabilization at the end of each epoch to stabilize the system (as described in Algorithm \ref{algo:ss}). However, as Figures \ref{fig:addition_after}--\ref{fig:mnist_after} show (only recurrent weight matrices are shown) this causes the network to become more brittle i.e., a tiny perturbation changes the spectral portrait significantly. Increasing the number of Power iterations does not alter the resulting contours of pseudo-spectrum, substantially. Thus, robustness and stability of neural networks although related, have to be addressed separately.

One can use spectral stabilization at the end of each epoch to stabilize the system (as described in Algorithm \ref{algo:ss}). However, as Figures \ref{fig:addition_after}--\ref{fig:mnist_after} show (only recurrent weight matrices are shown) this causes the network to become more brittle i.e., a tiny perturbation changes the spectral portrait significantly. Increasing the number of Power iterations does not alter the resulting contours of pseudo-spectrum, substantially. Thus, robustness and stability of neural networks although related, have to be addressed separately.

\begin{table}[]
\centering
\caption{\textbf{Accuracies of recurrent networks on three tasks.} }
\label{table:accuracy_rnn}
\tabcolsep=0.08cm
\resizebox{\columnwidth}{!}{%
\begin{tabular}{@{}llll@{}}
\toprule
Problem                          & RNN & LSTM & GRU \\ \midrule
Adding two numbers               &   99.73\%  &  99.87\%    &  99.94\%   \\
IMDB Sentiment analysis               &   83.29\%  &  83.81\%    &  85.96\%   \\
MNIST classification &  26.11\%   &  non-convergence    &  19.37-22.47\%   \\ \bottomrule
\end{tabular}
}
\end{table}

\begin{table}[]
\centering
\caption{\textbf{Degree of non-normality}: A large Henrici number is a necessary but not sufficient condition to detect spectral instability. LSTM and GRU have 4 and 3 gates, respectively.}
\label{table:nonnormal_rnn}
\tabcolsep=0.11cm
\resizebox{\columnwidth}{!}{%
\begin{tabular}{@{}llll@{}}
\toprule
Problem                          & RNN & LSTM & GRU \\ \midrule
Adding two numbers               &  1.6,2.3   & \makecell{5,3.6,1.3,4.5 \\ 4,3.5,2.2,4.2}    & \makecell{4.8,4.2,1.6 \\ 4,3.1,2.4}    \\
IMDB Sentiment analysis               &  0.2   & 2.6,3.8,1.2,2.4     & 0.9,1.8,0.8    \\
MNIST classification &  1.8   &  non-convergence    & 4.7,5.9,6    \\ \bottomrule
\end{tabular}
}
\end{table}

\section{Discussion and Open Problems}

It is well-known in the control community that simultaneously optimizing for robust stabilization and robust performance is a difficult problem. This paper reinforces this old adage by illustrating that recurrent networks that can appear to be highly accurate can be unstable. Furthermore, we show that learned matrices can indeed become non-normal, explaining the non-robustness of deep recurrent networks.

On one hand, a non-robust neural network is prone to adversarial attack \citep{Creswell2018}, whilst on the other hand, it exemplifies the brittleness of the weight matrices. Therefore, guaranteeing robustness of such networks in safety-critical scenarios is an absolute must. Stabilization using the spectrum is the first step but as the experiments show they do affect the robustness of the system. Stabilization in the control literature has an old history -- such as sensitivity minimization \citep{Zames1981}, gain-margin optimization \citep{Tannebaum1980}, etc. One way forward would be to use $H_{\infty}$ control to minimise the impact of a perturbation or to use $H_{\infty}$ loop-shaping to first achieve robust performance and then optimize the response to achieve robust stabilization. However, note that the spectral abscissa and the corresponding radius are non-convex functions; they are also non-Lipschitz (non-smooth) for polynomials that have multiple roots. Thus, a global minimization of either spectral abscissa or radius is non-trivial and goes far beyond constraining eigen-spectrum inside the stable unit disk. 

In summary, the balance between stability, robustness and generalisation is a crucial aspect of all optimisation, classification and inference problems. One perspective on this issue is to consider the optimisation of neural network weights as an inference or learning problem \citep{Sengupta2017}. On this view, the goal is to provide a parsimonious but accurate explanation for data features. Formally, this can be cast in terms of approximate Bayesian inference via maximisation of variational free energy or the evidence lower bound (ELBO). This may be an important perspective because generalisation necessarily entails minimising complexity \citep{Penny2010,Friston2011}. In other words, the entropy (of posterior beliefs about how the data were caused -- or what class generated them) should not only provide an accurate (low energy) account but should be sensitive to Occam`s principle in the sense of maximising the entropy of posterior beliefs (i.e., not committing to a particular posterior explanation). Crucially, to include entropy or complexity into the objective function, one has to accommodate uncertainty or beliefs about hidden states and parameters in the optimisation.

Most neural network approaches preclude this and focus on accuracy -- at the expense of complexity; thereby exposing themselves to overfitting and poor generalisation. An exception here would be variational autoencoders (VAE); provided both latent states and parameters were treated as random variables. It is therefore possible that generalised VAE may provide a principled solution to the stability/robustness trade-off – by implicitly finding solutions with the right sort of instability that retains a degree of robustness. The heuristic here rests upon the fact that, for timeseries data, one generally wants connection weights whose Jacobian has eigenvalues that approach the unit circle from within. In other words, optimal solutions are generally those that `remember' sequential information; such that the `memory' decays slowly over time. This takes the system to the edge of exponential divergence (i.e., a transcritical bifurcation) -- but only to the edge. When viewed like this, any neural network with deep memory in its internal dynamics is likely to (self) organise and exhibit critical slowing (i.e., self-organised criticality). 

It is fairly easy to show -- at least heuristically -- that systems that maximise variational free energy (e.g., variational auto-encoders) necessarily show this sort of critical slowing in response to perturbations -- such as new data features. Interestingly, the feature of the free energy functional that promotes instability is exactly the feature that minimises complexity -- and underwrites generalisation. In brief, this follows from the fact that the entropy of posterior beliefs about hidden variables corresponds to the curvature of the free energy functional at its maximum. This curvature can also be construed as scoring the brittleness of the problem at hand. In short, explicitly optimising an evidence bound (i.e., variational free energy) ensures generalisation, non-brittle solutions (with respect to inferred variables) and yet mandates critical slowing (from a dynamical perspective and associated instability). See \citet{Friston2012} for a discussion from the point of view of dynamical systems and variational inference. We conclude by suggesting the following open problems, answering them can lead to the usage of neural networks in safety-critical systems:

\begin{description}
\item[Open Problem 1] Design an algorithm that simultaneously optimizes performance (accuracy, etc.) and robustness (using pseudo-spectrum) while guaranteeing stability.
 \item[Open Problem 2] Could variational (e.g., variational message passing) neural networks dissolve the stability/robustness dilemma in a principled way?
\end{description}



\bibliography{rnn}
\bibliographystyle{icml2018}

\end{document}